\documentclass[letterpaper, 10 pt, conference]{ieeeconf}
\IEEEoverridecommandlockouts
\title{\LARGE \bf
Learning Robust Execution in Robotic Manipulation with Agentic Reinforcement Learning
}

\author{Xiaopeng Zhang$^{1}$, Yueyang Weng$^{1}$, Qi Liu$^{2}$, Yongjin Mu$^{1}$ and Yanjie Li$^{1}$$^\dagger$
\thanks{This work was supported by Shenzhen Fundamental Research ZDCY20250901095402003, Guangdong Provincial Natural Science Foundation 2026A1515011727 and the Open Fund of Innovation Center for Control Actuators.}
\thanks{$^{1}$ School of Inteligence Science and Engineering, the Harbin Institute of Technology Shenzhen, 518055, China. {\tt\small  autolyj@hit.edu.cn}}
\thanks{$^{2}$ Faculty of Robot Science and Engineering, Northeastern University, Shenyang, 110819, China.}
\thanks{$^\dagger$ Corresponding author}
}

\usepackage[T1]{fontenc}
\usepackage{amsmath}
\usepackage{amssymb}
\usepackage{graphicx}
\usepackage{booktabs}
\usepackage{algorithm}
\usepackage{algpseudocode}
\usepackage{multirow}
\usepackage[table]{xcolor}
\usepackage{cite}    
\makeatletter
\let\NAT@parse\undefined
\makeatother
\usepackage[hidelinks]{hyperref}
\begin{document}

\maketitle
 \thispagestyle{empty}
\pagestyle{empty}

\begin{abstract}
Robotic manipulation poses fundamental challenges due to uncertainty, long-horizon execution, and compounding errors, which can easily destabilize execution and lead to task failure. Although recent vision-language-action (VLA) models exhibit strong generalization, they typically lack explicit mechanisms to assess execution stability and to recover when execution deviates from its nominal behavior. In this paper, we propose: (1) two complementary metrics to assess execution quality at runtime, and (2) an agentic reinforcement learning framework that learns to restore effective execution through high-level decision-making rather than directly learning low-level actions. In this framework, an agentic policy reasons over recent execution history and selects among a small set of execution modes to regulate the execution process. Under execution degradation, it triggers appropriate recovery mechanisms to restore the robot to previously visited nominal states, enabling the task to continue. We evaluate the proposed method on the LIBERO benchmark, achieving up to a 13.7\% improvement in success rate under standard settings and up to a 39.2\% improvement under disturbance settings, demonstrating substantially enhanced execution robustness.

\label{abs}
\end{abstract}

\section{INTRODUCTION}
\label{intro}
Robotic manipulation is a fundamental capability for embodied intelligent systems \cite{bai2025towards,bai2025embodied,kawaharazuka2025vision}, yet remains challenging due to uncertainty, contact-rich interactions, and long-horizon execution. Even when a manipulation policy performs well under nominal conditions, small deviations in perception and dynamics can accumulate over time, causing execution to drift away from expected behavior. Such execution degradation often leads to unstable behaviors, loss of task progress, or irreversible failures, directly limiting robustness in complex manipulation tasks \cite{11245856,zhang2025robustvla}.

\begin{figure}[t]
    \centering
    \includegraphics[width=\columnwidth]{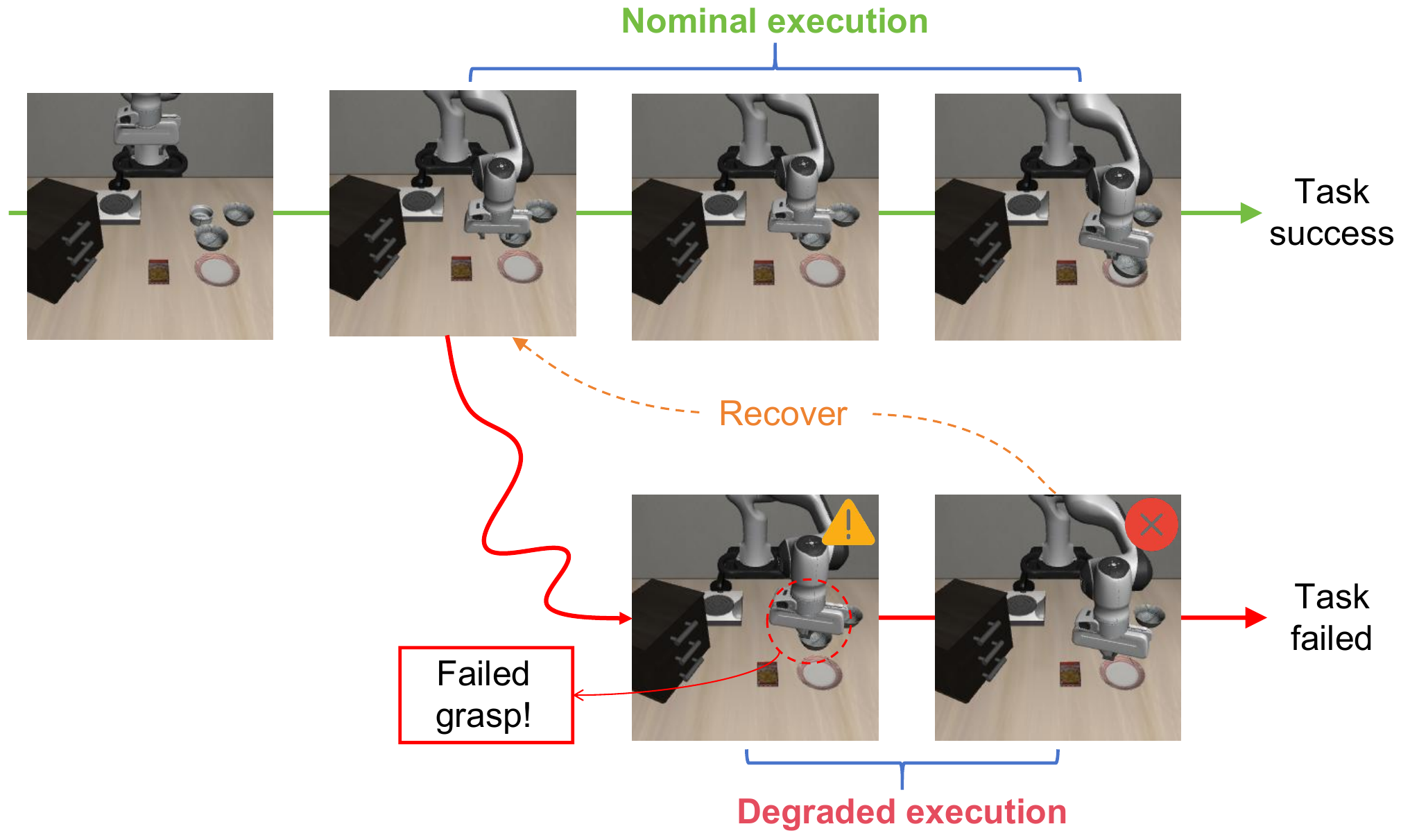}
    \caption{Under nominal conditions, manipulation policies can successfully complete tasks through stable execution. However, small execution errors may accumulate over time, causing execution to drift into degraded states. Once execution becomes unstable, task progress stalls and failures occur, motivating explicit mechanisms for execution-level monitoring and recovery.}
    \label{fig:background}
    \vspace{-10pt}
\end{figure}

Recent progress in learning-based manipulation policy, particularly vision-language-action (VLA) models, demonstrates strong generalization across objects, tasks, and scenes through large-scale pretraining \cite{kim2024openvla,black2024pi_0,intelligence2025pi_}. However, despite their impressive performance under familiar distributions, VLAs remain vulnerable to execution-level deviations during deployment in long-horizon tasks. As illustrated in Fig.~\ref{fig:background}, execution can bifurcate from a nominal trajectory due to slight grasp misalignment. Without active recovery, the system drifts into a degraded state where the policy is no longer effective, causing small errors to accumulate into irreversible task failures. We refer to such failures, where execution deviates from nominal behavior, as execution-level failures.

One straightforward way to mitigate such failures is to expand training data and further fine-tune VLAs to cover a broader range of conditions \cite{kim2024openvla}. Although effective, this approach is often resource-intensive due to the high cost of data collection and the substantial computational overhead required to retrain foundation-scale models. Moreover, the long-tail distribution of failure cases limits the effectiveness of data scaling alone. As a result, relying solely on data scaling can be costly and may not always suffice.

An alternative strategy introduces additional reasoning components, such as vision-language models (VLMs), to detect failures and replan execution \cite{yang2025agentic,duan2024aha}. Although these methods demonstrate strong recovery and replanning capabilities, they often introduce additional system complexity and inference latency, which can limit scalability and efficiency. Moreover, language-driven replanning is sensitive to instruction distribution shifts, further complicating reliable execution management \cite{11246863}.

Notably, execution-level failures are not unique to VLAs. Similar issues arise in manipulation methods trained via imitation learning \cite{chi2025diffusion}, where policies lack explicit mechanisms to assess execution quality or adapt when execution deviates from expected behavior.

To address these issues, we improve execution robustness from an execution-level perspective. First, we design two complementary metrics, termed local execution quality and global execution quality, to assess execution performance at runtime. Second, instead of learning new actions or relying on heavy external reasoning modules, we regulate the execution process of a manipulation policy. We formulate this as an \textbf{agentic reinforcement learning} problem by proposing a high-level agentic policy that observes recent execution history and selects among a set of execution modes (e.g., execute, retry, repair, reset) to determine how execution proceeds. Importantly, the agentic policy does not generate control commands but governs the application of the low-level policy. We term this formulation \textit{agentic} because the learned policy acts as an execution manager, separating execution regulation from action generation. In this way, the overall system evolves from a purely conditional action generator into a decision-driven execution framework with exploration and recovery capability.

We apply the proposed method to multiple VLAs and diffusion policy, and evaluate on the LIBERO benchmark \cite{liu2023libero}, covering four subsets: \textit{LIBERO-Spatial}, \textit{LIBERO-Object}, \textit{LIBERO-Goal}, and \textit{LIBERO-Long}. Experimental results show that our method consistently improves task success rates and enhances recovery from execution degradation without modifying or retraining the low-level policies. Moreover, it demonstrates improved robustness under disturbance.

\begin{figure*}[htbp]
    \centering
    \includegraphics[width=0.9\textwidth]{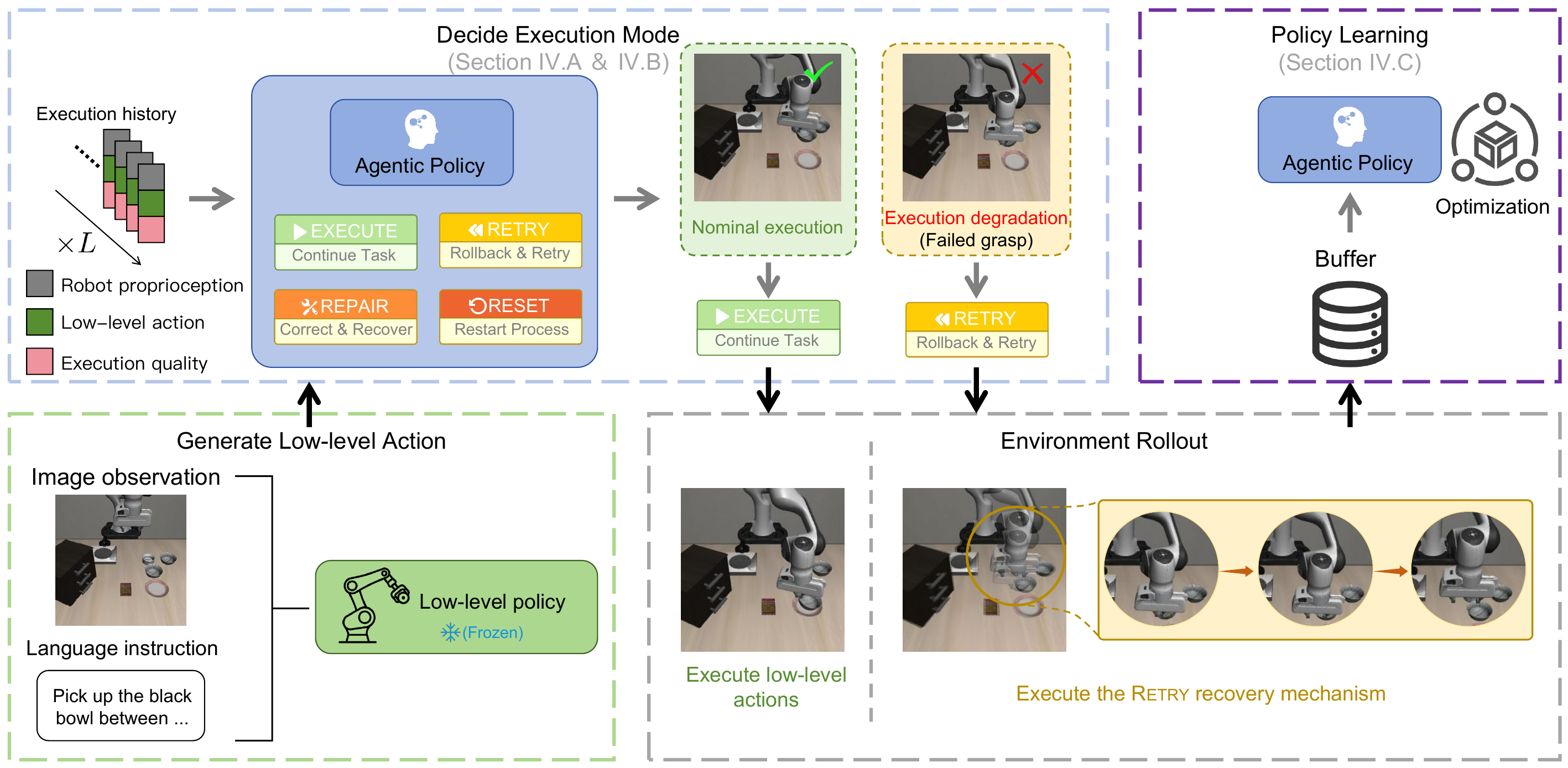} 
    \caption{Agentic reinforcement learning framework. A frozen low-level policy generates low-level actions, while a high-level agentic policy observes execution history and selects execution modes (\textsc{Execute}, \textsc{Retry}, \textsc{Repair}, \textsc{Reset}) to regulate the execution process. Rollout data are stored for policy learning, enabling recovery from execution degradation without modifying the low-level policy.}
    \label{fig:framework}
    \vspace{-10pt}
\end{figure*}

\section{Related Work}
\label{related_work}
\subsection{Vision-Language-Action Models for Robotic Manipulation}
Vision-language-action models have emerged as a powerful paradigm for robotic manipulation by leveraging large-scale pretraining on diverse multimodal datasets \cite{kim2024openvla,black2024pi_0,intelligence2025pi_,zitkovich2023rt}. By directly mapping visual observations and language instructions to actions, these methods demonstrate strong generalization across objects, tasks, and scenes, and have advanced the scalability of learning-based manipulation systems. Recent benchmarks and real-world deployments show that VLAs can successfully execute complex, multi-step manipulation tasks under nominal conditions \cite{intelligence2025pi,fan2025long}. 

Despite these advances, VLAs are typically deployed as monolithic executors and lack explicit mechanisms to monitor execution quality or adapt behavior once execution deviates from its nominal trajectory. During long-horizon tasks, small disturbances, contact uncertainty, and compounding errors can gradually degrade execution, leading to unstable behavior and task failure \cite{yang2025seqvla,din2025vision,11246863}.

Our method complements VLAs by proposing a lightweight, execution-level decision mechanism that operates on top of frozen manipulation policies. Rather than modifying or retraining VLAs, we focus on learning to manage execution when degradation occurs to improve robustness.

\subsection{Failure Recovery and Hierarchical Decision-Making}
Failure recovery in robotic manipulation has evolved from reactive correction strategies to more structured hierarchical decision-making frameworks \cite{li2026failure,9981315,yang2025agentic}. Early approaches rely on explicit failure detection and local corrective behaviors. Runtime monitoring methods such as \cite{AgiaSinhaEtAl2024} analyze execution consistency to detect policy deviations, and \cite{romer2025fiper} estimate the likelihood of execution failure at runtime, but they do not learn recovery behaviors themselves.

Hierarchical reinforcement learning (HRL)~\cite{pateria2021hierarchical} employs structured decision layers above low-level controllers, where a high-level policy selects skills executed by a lower-level policy. For example, \cite{11245856} models recovery as a hierarchical RL problem, where a recovery policy operates over primitive actions and nominal controllers to return the system to states from which nominal execution can succeed.

Recent research explores dual-system architectures, which introduce explicit high-level reasoning above reactive control, separating fast visuomotor execution from slower semantic planning and verification. These methods typically introduce an additional VLM as a reasoning component to interpret task progress and guide recovery. \cite{yang2025agentic} exemplifies this paradigm by combining a reasoning model for subgoal decomposition, a VLA executor, and a VLM-based verifier, forming a closed-loop coordination mechanism.

In contrast, our method focuses on execution-level regulation of a low-level policy. Our method learns a lightweight agentic policy that selects among predefined execution modes to regulate how execution proceeds, including when to invoke recovery. Unlike dual-system methods that rely on semantic reasoning or verification to guide recovery, the agentic policy operates purely on execution history and execution quality signals to make low-latency execution decisions.

\section{Problem Statement}
\label{section:problem_statement}
We consider the execution degradation in robotic manipulation, where a robot executes a task using a frozen low-level policy (e.g., a VLA policy), but disturbances, contact uncertainty, and compounding errors cause the robot to gradually deviate from nominal behavior.

To address this issue, we propose an agentic reinforcement learning (RL) framework, as shown in Fig.~\ref{fig:framework}. In this framework, a frozen low-level policy continuously outputs low-level actions, while a high-level agentic policy regulates the execution process. Instead of generating low-level control actions directly, the agentic policy decides whether to proceed with the low-level policy or to trigger predefined recovery mechanisms based on execution history. We formulate this process as a partially observable Markov decision process (POMDP), defined by tuple $\mathcal{M} = \langle \mathcal{S}, \mathcal{A}, P, R, \gamma \rangle$. Here, $\mathcal{S}$ represents the state space, which is partially observable. $\mathcal{A}$ denotes a discrete set of agentic execution modes. $P:\mathcal{S} \times \mathcal{A} \times \mathcal{S} \rightarrow [0,1]$ describes the transition dynamics determined jointly by the simulation environment and the low-level policy. $R:\mathcal{S} \times \mathcal{A} \rightarrow \mathbb{R}$ is the reward function, and $\gamma$ is the discount factor.

Given a low-level policy $\hat{\pi}$ and a high-level agentic policy $\pi_{\text{agent}}$, at each time step $t$, $\hat{\pi}$ continuously generates low-level actions $\hat{a}_t$ conditioned on visual observations and language instructions. In contrast, the agentic policy $\pi_{\text{agent}}$ runs at decision steps with a fixed time interval.

At each decision step, $\pi_{\text{agent}}$ observes an execution history $h_t$ (comprising recent proprioception, low-level actions, and execution quality) with a fixed horizon and selects an execution mode $a_t \in \mathcal{A}$ to regulate the execution process. The action space is defined as a finite set: $\mathcal{A} = \{ \textsc{Execute}, \textsc{Retry}, \textsc{Repair}, \textsc{Reset} \}$. Specifically, \textsc{Execute} mode continues executing the actions output by $\hat{\pi}$, and other modes trigger distinct recovery mechanisms to handle execution degradation. The objective of the proposed framework is to learn $\pi_{\text{agent}}$ via RL to enhance execution robustness, ensuring the robot can recover from execution degradation and successfully complete tasks.

\begin{figure}[t]
    \centering
    \includegraphics[width=\columnwidth]{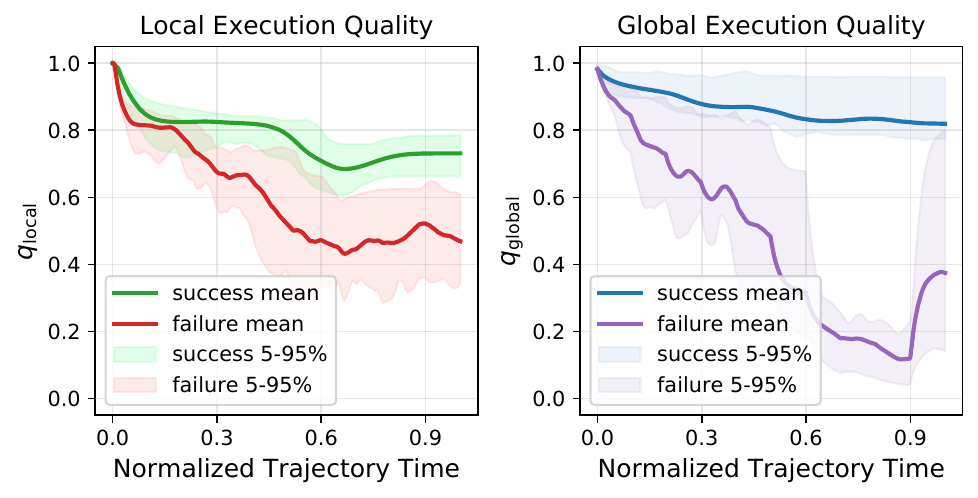}
    \caption{Local and global execution quality over normalized trajectory time for successful and failed executions. Successful trajectories maintain higher and more stable execution quality, while failed trajectories exhibit faster degradation. Shaded regions indicate the 5th–95th percentile ranges.}
    \label{fig:local_exec_quality}
    \vspace{-10pt}
\end{figure}

We refer to $\pi_{\text{agent}}$ as the \emph{agentic policy}, and the overall formulation as \emph{agentic reinforcement learning}, because the policy operates as an autonomous execution-level decision maker. Rather than generating low-level control actions, it governs how the low-level policy $\hat{\pi}$ is applied over time. This explicitly separates execution management from action generation. For notational convenience, we denote the agentic policy as $\pi_\theta$ in the remainder of the paper, where $\theta$ represents the network parameters.

\begin{figure*}[htbp]
    \centering
    \includegraphics[width=0.9\textwidth]{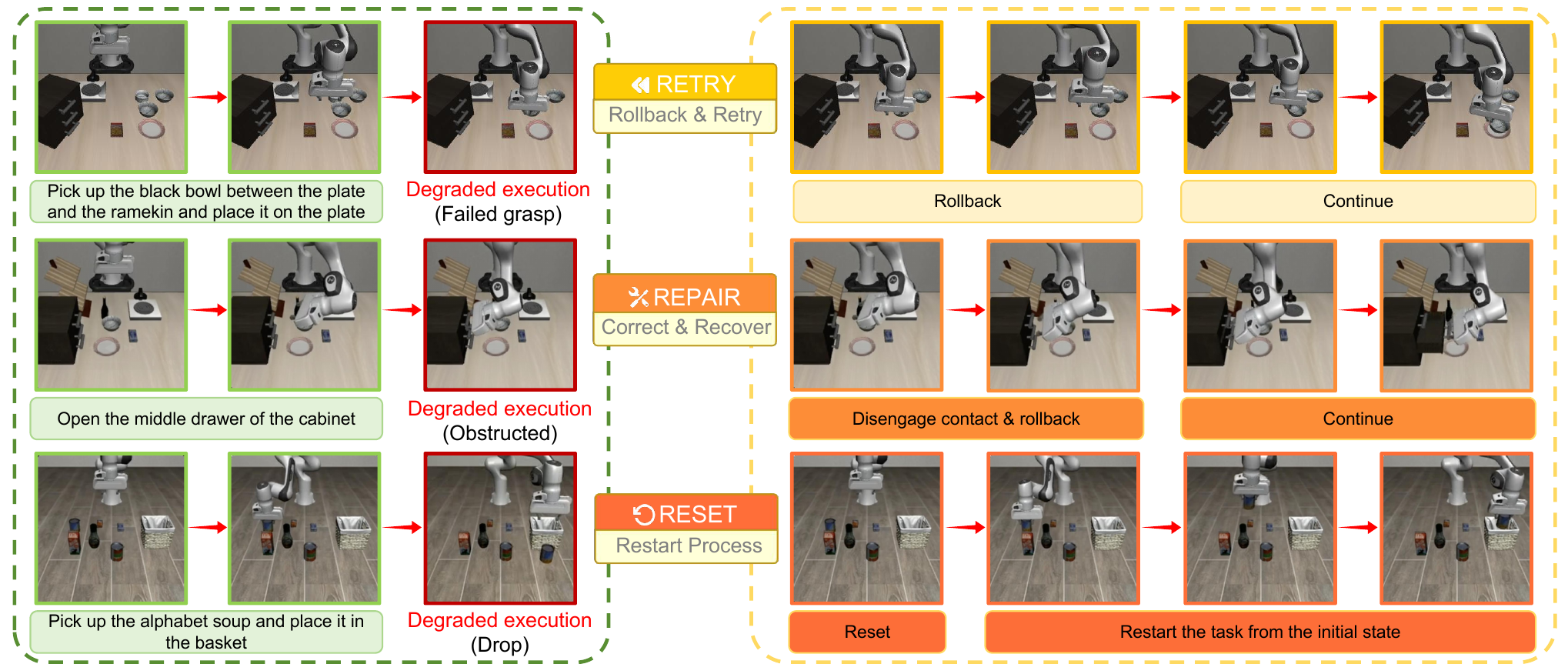}
    \caption{Illustration of the three recovery mechanisms under different execution degradation cases. \textsc{Retry} restores the robot to a recent high-quality state and resumes execution. \textsc{Repair} restores the robot to a contact-free high-quality state within a longer horizon before continuing execution. \textsc{Reset} terminates the current episode and restarts the task from the initial configuration.}
    \label{fig:exe_modes}
    \vspace{-12pt}
\end{figure*}

\section{Method}
\label{method}

\subsection{Local and Global Execution Quality}
\label{subsection:exec_quality}
Under nominal conditions, a policy should complete tasks smoothly, and the trajectories should exhibit consistent behavioral patterns. To assess execution quality, we propose two complementary metrics. \textbf{Local execution quality} monitors short-term stability, while \textbf{global execution quality} captures long-horizon drift from nominal behaviors. Together, these metrics summarize temporal execution behavior and inform the decision-making of the agentic policy.

\subsubsection{Local Execution Quality}
At each step $t$, we maintain a sliding window of size $W$ over recent end-effector positions $\{p_{k}\}_{k=t-W+1}^{t}$ and low-level actions $\{\hat{a}_{k}\}_{k=t-W+1}^{t}$.

We first compute a \textit{motion effectiveness} term $E$ to detect stagnation or jamming. Intuitively, this metric measures the ratio between the actual distance moved by the end-effector and the commanded action magnitude. A low $E$ indicates that the robot is exerting control effort but making little physical progress:
\begin{equation}
    E = \frac{\lVert p_t - p_{t-W+1} \rVert}{\frac{1}{W}\sum_{k=t-W+1}^{t} \lVert \hat{a}_k \rVert + \epsilon}.
\label{eq:effectiveness}
\end{equation}
Here, $W$ is the window size and $\epsilon$ ensures numerical stability. To map this unbounded ratio to a normalized range $[0, 1]$, we apply a saturation transform:
\begin{equation}
    E_{\text{norm}} = \frac{E}{E + c}.
\end{equation}
The scalar $c$ controls the sensitivity of the saturation, ensuring that the metric smoothly approaches $1$ as effectiveness increases.

Simultaneously, we assess \textit{motion smoothness} $S$ based on the instantaneous speeds $v_k = \lVert p_k - p_{k-1} \rVert$. We utilize the inverse coefficient of variation to penalize erratic fluctuations:
\begin{equation}
    S = \frac{1}{1 + \frac{\sigma_v^2}{\mu_v^2 + \epsilon}},
\end{equation}
where $\mu_v$ and $\sigma_v^2$ denote the mean and variance of the speeds $\{v_k\}$ within the sliding window, respectively.

Subsequently, we aggregate these terms via a weighted sum, normalize the result through a sigmoid function $\sigma(\cdot)$, and finally apply exponential moving average smoothing:
\begin{equation}
    \begin{split}
        q_{\text{local}}(t) &= \beta q_{\text{local}}(t-1) \\
        &\quad + (1-\beta) \cdot \sigma\left( k \cdot (w_1 E_{\text{norm}} + w_2 S - b) \right).
    \end{split}
\end{equation}
Here, $b$ and $k$ determine the threshold and sensitivity, and $\beta$ controls smoothing. The local execution quality score is initialized to $q_{\text{local}}(0)=1.0$ at the beginning of each episode.

\subsubsection{Global Execution Quality}

While local quality measures physical stability, it fails to detect execution-level deviation (e.g., moving smoothly but towards an incorrect target). To address this, we propose a global execution quality $q_{\text{global}}$ that compares the current execution prefix against successful reference trajectories.

At each time step $t$, we encode the recent execution window into a compact feature vector $z_t = [\mu_s, \mu_{\hat{a}}, m, \bar{a}]$, where $\mu_s$ and $\mu_{\hat{a}}$ denote the window-averaged robot state (including end-effector pose and gripper width) and low-level action, $m = \lVert p_t - p_{t-W+1} \rVert$ is the net displacement, and $\bar{a} = \frac{1}{W}\sum_{k=t-W+1}^{t} \lVert \hat{a}_k \rVert$ is the mean action magnitude.

Before deployment, we collect $N=50$ successful trajectories for each task using the frozen low-level policy. For each trajectory of max length $T$, we compute the feature vector $z_t$ at every time step $t$. To account for temporal misalignment across trajectories of different lengths, we normalize each time step by its progress ratio $\frac{t}{T}$ and discretize the normalized progress into $B=10$ bins. Feature vectors whose normalized progress falls into the same bin are grouped together, forming a stage-aware reference library. For a new task, this reference library can be extended by collecting additional successful trajectories, enabling the global execution quality to be readily applied.

During execution, the current feature $z_t$ is compared against the $k$ nearest neighbors ($k=5$) within the corresponding progress bin of the reference library. Let $d_t$ denote the average Euclidean distance to these neighbors. The global quality score is then computed as $q_{\text{global}}(t) = \beta q_{\text{global}}(t-1) + (1-\beta)\exp(-\alpha d_t)$, where $\alpha$ controls sensitivity to deviation. A high score indicates close alignment with successful behaviors, whereas persistent deviation leads to a lower score. The global execution quality score is initialized to $q_{\text{global}}(0)=1.0$ at the beginning of each episode.

As shown in Fig.~\ref{fig:local_exec_quality}, successful executions maintain higher and more stable quality, while failed executions exhibit rapid local instability and pronounced long-horizon degradation.

\subsection{Recovery Mechanisms}
\label{subsection:recovery_mechanisms}
To restore the robot to nominal execution when execution degradation occurs, we design three recovery mechanisms: \textsc{Retry}, \textsc{Repair}, and \textsc{Reset}, which correspond to the latter three actions in the agentic action space defined in Section~\ref{section:problem_statement}. These three mechanisms are designed to cover the majority of recovery behaviors required under common execution degradation scenarios. 

Our key insight is that different degrees of execution degradation require different levels of recovery. For instance, a slight grasp misalignment may only require a short rollback, whereas more severe contact misalignment may require temporarily releasing contact and repositioning the robot before continuing. Importantly, these recovery mechanisms are not intended to generate new actions, but to restore the robot to previously visited nominal states, thereby recovering the effectiveness of the low-level policy.

First, we aggregate the local and global execution quality scores proposed in Section~\ref{subsection:exec_quality} into a single scalar score at each time step $t$:
\begin{equation}
q_{\text{agg}}(t) = \lambda \, q_{\text{local}}(t) + (1 - \lambda)\, q_{\text{global}}(t).
\end{equation}
Here, $\lambda \in [0,1]$ is a weighting coefficient. This aggregated score is used to evaluate overall execution health and serves as a criterion for selecting rollback states during recovery.

In practice, both \textsc{Retry} and \textsc{Repair} realize rollback by using an operational space controller (OSC) to drive the robot back to a selected historical state. During this process, the gripper is commanded to remain open, and the OSC tracks incremental end-effector pose commands as control targets. In contrast, \textsc{Reset} handles irrecoverable severe errors by terminating the current episode and restarting the task from the initial configuration.
\textsc{Retry} drives the robot back to the state with the highest aggregated execution quality $q_{\text{agg}}$ among the most recent $M$ steps and then resumes nominal execution. 
\textsc{Repair} performs a stronger rollback by restoring the robot to the state with the highest $q_{\text{agg}}$ among the contact-free states within the most recent $N$ steps $(N>M)$. Contact-free states are identified by thresholding the end-effector contact force from the MuJoCo simulator~\cite{6386109}. Contact forces are not used as continuous inputs. Instead, we compute a binary contact indicator using a fixed threshold $\tau=5\,\mathrm{N}$.  This treatment provides broad generality and can be applied to different contact signals, such as force sensors and joint torques.

While the rollback horizons $M$ and $N$ can be learned, we find that this reduces training stability. Fixed values $M=15$ and $N=30$ are sufficient and used throughout. Fig.~\ref{fig:exe_modes} illustrates the recovery mechanisms triggered by the agentic policy under different types of execution degradation.

\subsection{Agentic Reinforcement Learning}
\label{subsection:agentic_rl}
Our goal is to learn an agentic policy that determines how execution should proceed based on recent execution history, including proprioception, low-level actions, and execution quality.
Specifically, when execution progresses normally, the agentic policy allows the robot to continue following the low-level policy (\textsc{Execute}). When encountering execution degradation, the agentic policy adaptively selects an appropriate recovery mechanism (\textsc{Retry}, \textsc{Repair}, \textsc{Reset}).

\subsubsection{Observation}
At each decision step, the agentic policy observes a fixed-length execution history $h_t$, constructed as a sliding window over the most recent $L$ low-level steps. The history includes recent robot proprioception, low-level actions produced by $\hat{\pi}$, and the two execution quality scores. This compact representation summarizes execution dynamics and progress without requiring access to raw visual observations or privileged simulator states.

The critic has access to additional privileged global state information from the simulator, which is not available to the actor at test time. This asymmetry improves value estimation under sparse and delayed rewards while preventing information leakage to the deployed policy.

Notably, information generated during recovery phases is not recorded into the execution history. This prevents recovery behaviors from contaminating the history used for subsequent decision making.

\subsubsection{Action Space}
The agentic policy makes decisions at an interval of $K$ low-level steps. At each agentic decision step, the agentic policy outputs a categorical distribution over execution modes
\begin{equation}
    a_t \in \mathcal{A} = \{ \textsc{Execute}, \textsc{Retry}, \textsc{Repair}, \textsc{Reset} \}.
\end{equation}
The selected action determines how execution proceeds in the subsequent phase. \textsc{Execute} allows the low-level policy to continue nominal execution until the next decision step, while the other modes invoke predefined recovery mechanisms. The agentic policy does not generate low-level control commands. In implementation, the action space is encoded as four discrete integers $\{0,1,2,3\}$ corresponding to the four execution modes. During training, the critic outputs a scalar value used to evaluate the expected return. 

\subsubsection{Reward Function and Optimization Objective}
We design the reward function to encourage task completion and efficient execution while penalizing unnecessary recovery. Specifically, task success and failure yield terminal rewards of $+1.0$ and $-1.0$ respectively, and each timestep incurs a small time penalty of $-0.02$ to discourage prolonged execution. Invoking recovery mechanisms introduces fixed costs proportional to their recovery strength, with penalties of $-0.1$, $-0.3$, and $-0.5$ for \textsc{Retry}, \textsc{Repair}, and \textsc{Reset}, respectively. \textsc{Execute} incurs no additional cost. The complete reward components are summarized in Table~\ref{tab:training_details} (left).

The agentic policy $\pi_\theta$ is trained to maximize the expected discounted return
\begin{equation}
    \max_{\pi_\theta} \; \mathbb{E}_{\pi_\theta} \left[ \sum_{t=0}^{T} \gamma^t r_t \right],
\end{equation}
where $r_t$ denotes the reward at time step $t$ and $\gamma$ is the discount factor. We optimize $\theta$ using Proximal Policy Optimization (PPO)~\cite{schulman2017proximal}, which updates the policy by maximizing a clipped surrogate objective to ensure stable learning in the discrete action space.
Both the agentic policy and the critic are implemented as lightweight multi-layer perceptrons. Table~\ref{tab:training_details} (right) summarizes the main training hyperparameters.

\begin{table}[t]
    \centering
    \caption{Reward components and training hyperparameters.}
    \label{tab:training_details}
    \vspace{0.3em}
    \renewcommand{\arraystretch}{1.1}
    \setlength{\tabcolsep}{6pt}
    \small
    \begin{tabular}{l c | l c}
        \toprule
        \multicolumn{2}{c|}{\textbf{Reward}} &
        \multicolumn{2}{c}{\textbf{Training Hyperparameters}} \\
        \midrule
        Success reward      & $+1.0$  & Learning rate     & $1\!\times\!10^{-4}$ \\
        Failure penalty     & $-1.0$  & Discount $\gamma$ & $0.99$ \\
        Time penalty        & $-0.02$ & PPO clip          & $0.2$ \\
        \textsc{Execute} cost        & $0$     & Value coef.       & $0.1$ \\
        \textsc{Retry} cost          & $-0.1$  & Entropy coef.     & $0.01$ \\
        \textsc{Repair} cost         & $-0.3$  & Decision interval $K$ & $5$ \\
        \textsc{Reset} cost          & $-0.5$  & History length $L$    & $20$ \\
        \bottomrule
    \end{tabular}
    \vspace{-8pt}
\end{table}

\begin{table*}[htbp]
    \centering
    \caption{Quantitative Results on the LIBERO Benchmark. We report success rates (\%) under two settings: (I) Standard nominal conditions, and (II) Robustness testing with random kinematic disturbances. Shaded rows highlight the average performance gains ($\Delta$), demonstrating significant resilience in the disturbance setting.}
    \label{tab:libero_main}
    \vspace{0.5em}
    \renewcommand{\arraystretch}{1.2}
    \setlength{\tabcolsep}{6pt}
    \resizebox{0.9\textwidth}{!}{\begin{tabular}{lcccccccc}
        \toprule
        \multirow{2}{*}{\textbf{Method}}
        & \multicolumn{2}{c}{\textbf{Spatial}} 
        & \multicolumn{2}{c}{\textbf{Object}} 
        & \multicolumn{2}{c}{\textbf{Goal}} 
        & \multicolumn{2}{c}{\textbf{Long}} \\
        \cmidrule(lr){2-3}
        \cmidrule(lr){4-5}
        \cmidrule(lr){6-7}
        \cmidrule(lr){8-9}
        & w/o Agentic & w/ Agentic
        & w/o Agentic & w/ Agentic
        & w/o Agentic & w/ Agentic
        & w/o Agentic & w/ Agentic \\
        
        \midrule
        \multicolumn{9}{l}{\textit{\textbf{I. Standard Evaluation (Nominal Conditions)}}} \\
        \midrule
        OpenVLA (FT)
            & 77.8 & \textbf{90.2}
            & 70.8 & \textbf{88.7} 
            & 74.0 & \textbf{92.4} 
            & 54.0 & \textbf{74.5}  \\
        $\pi_0$ (FT)
            &   96.4   & \textbf{97.2}
            &   97.0   & \textbf{97.4}
            &   \textbf{96.8}   & 96.2
            &   81.0   & \textbf{90.6} \\
        $\pi_{0.5}$ (FT)
            &   \textbf{97.4}   & 96.6
            &   97.4   & \textbf{98.2}
            &   \textbf{98.0}   & 97.4
            &   92.4   &  \textbf{95.2} \\
        Diffusion Policy
            &   78.3   & \textbf{86.4}
            &   90.2   & \textbf{92.5}
            &   68.3   & \textbf{77.6}
            &   50.5   &  \textbf{72.4} \\
        \midrule

        \rowcolor{gray!10}
        \textbf{Avg. $\Delta$}
            & \multicolumn{2}{c}{\textbf{+5.1}}
            & \multicolumn{2}{c}{\textbf{+5.4}}
            & \multicolumn{2}{c}{\textbf{+6.6}}
            & \multicolumn{2}{c}{\textbf{+13.7}} \\

        \bottomrule
        \noalign{\vspace{2ex}}
        \toprule
        
        \multicolumn{9}{l}{\textit{\textbf{II. Robustness Evaluation (w/ Random Disturbance)}}} \\
        \midrule
        OpenVLA (FT)
            & 47.2 & \textbf{83.0}
            & 43.6 & \textbf{76.2} 
            & 47.0 & \textbf{81.4} 
            & 33.4 & \textbf{67.6}  \\
        $\pi_0$ (FT)
            & 58.0 & \textbf{85.5}
            & 60.0 & \textbf{86.0}
            & 58.0 & \textbf{85.0}
            & 33.8 & \textbf{79.5} \\
        $\pi_{0.5}$ (FT)
            & 80.0 & \textbf{90.4}
            & 64.0 & \textbf{92.8}
            & 67.2 & \textbf{88.8}
            & 40.2 & \textbf{87.2} \\
        Diffusion Policy
            & 45.0 & \textbf{74.0}
            & 48.2 & \textbf{70.5}
            & 45.0 & \textbf{75.0}
            & 30.5 & \textbf{60.5} \\
        \midrule

        \rowcolor{gray!10}
        \textbf{Avg. $\Delta$}
            & \multicolumn{2}{c}{\textbf{+25.7}}
            & \multicolumn{2}{c}{\textbf{+27.4}}
            & \multicolumn{2}{c}{\textbf{+28.3}}
            & \multicolumn{2}{c}{\textbf{+39.2}} \\
        \bottomrule
    \end{tabular}}
    \vspace{-8pt}
\end{table*}

\section{Experiments}
The experiments are designed to evaluate whether our method improves task success and execution robustness while keeping the low-level policy frozen.

In particular, we seek to answer the following questions: (1) Can the agentic policy improve success and robustness under execution degradation? (2) Can the framework be applied to different tasks and low-level policies? (3) Does the policy adapt decisions to degradation severity? (4) Do different decisions yield distinct execution quality outcomes? (5) Is robustness gained with limited cost?

\subsection{Experimental Setup}
\textbf{Benchmarks and Baselines.}
We evaluate on the four task suites of the LIBERO benchmark, namely \textit{LIBERO-Spatial}, \textit{LIBERO-Object}, \textit{LIBERO-Goal}, and \textit{LIBERO-Long}, which span increasing task diversity and long-horizon compositionality. We compare against representative end-to-end manipulation policies without execution-level decision-making, including OpenVLA, $\pi_0$, $\pi_{0.5}$, and diffusion policy, all initialized from models fine-tuned on LIBERO.

\textbf{Training and Evaluation.}
The agentic policy is trained separately for each task, with up to 1,000,000 high-level decision steps per task. Here, a training step corresponds to one agentic decision per environment, and the total step count is accumulated across all environments. Training spans multiple tasks within each LIBERO subset, and episodes terminate upon success or timeout.

At evaluation time, the learned agentic policy is deployed without fine-tuning and performance is measured by task \textbf{success rate} under identical task settings. When computing success rates, episodes in which \textsc{Reset} is invoked are counted as failures to ensure fair comparison.

To evaluate robustness, we additionally conduct a stress test with random disturbances. In each episode, a disturbance is injected once at a random timestep by replacing the low-level action with noise $\boldsymbol{\xi} \sim \mathcal{U}(-\delta, \delta)$ for $5$ consecutive steps, with $\delta=3.0$ in all experiments.

\subsection{Main Results (Q1, Q2)}
Table~\ref{tab:libero_main} reports task success rates. Our framework consistently improves performance for policies prone to execution degradation, with the largest gains observed for OpenVLA and the diffusion policy, especially on the challenging \textit{LIBERO-Long} suite (e.g., +20.5\% for OpenVLA). For strong baselines such as $\pi_0$ and $\pi_{0.5}$ that already achieve near-saturated performance, our method preserves their high success rates without degradation. Overall, the average improvement increases with task complexity, indicating effective mitigation of long-horizon execution degradation.

Under the disturbance setting (Table~\ref{tab:libero_main}, Part~II), baseline performance degrades substantially, most notably for OpenVLA, whose success rate drops to 33.4\% on \textit{LIBERO-Long}. In contrast, our method maintains a success rate of 67.6\%, demonstrating strong robustness to unmodeled state deviations. Across all baselines, the agentic policy consistently recovers from disturbed executions by triggering appropriate recovery routines, converting many otherwise irreversible failures into successful task completions.

\subsection{Further Analysis}
\subsubsection{Training Performance}
\begin{figure}[t]
    \centering
    \includegraphics[width=\columnwidth]{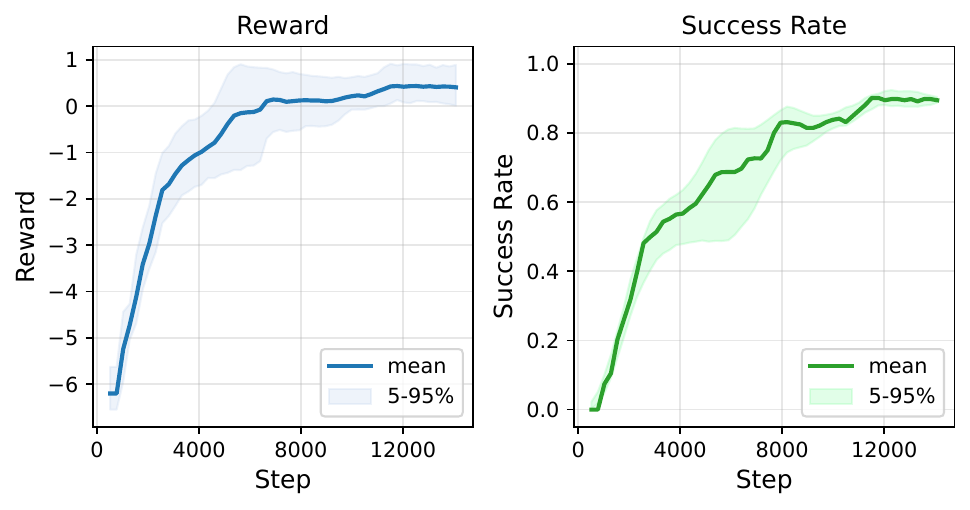}
    \caption{PPO training curves of the agentic policy on LIBERO-Spatial with a frozen OpenVLA policy. Mean reward (left) and success rate (right) are shown with 95\% CIs over 3 seeds.}
    \label{fig:training_plot}
    \vspace{-10pt}
\end{figure}
Fig.~\ref{fig:training_plot} presents the PPO training curves of the agentic policy trained in the \textit{LIBERO-Spatial} environment with a frozen OpenVLA policy. Both the training reward and task success rate increase steadily over training, with shaded regions indicating 95\% confidence intervals (CIs) across three random seeds.

The smooth and consistent improvement in both metrics suggests that the agentic policy can be trained stably using sparse task-level rewards, and that execution-level decisions learned through PPO effectively improve task outcomes without modifying the low-level policy.

\subsubsection{Decision Distribution (Q3)}
\begin{figure}[t]
    \centering
    \includegraphics[width=0.87\columnwidth]{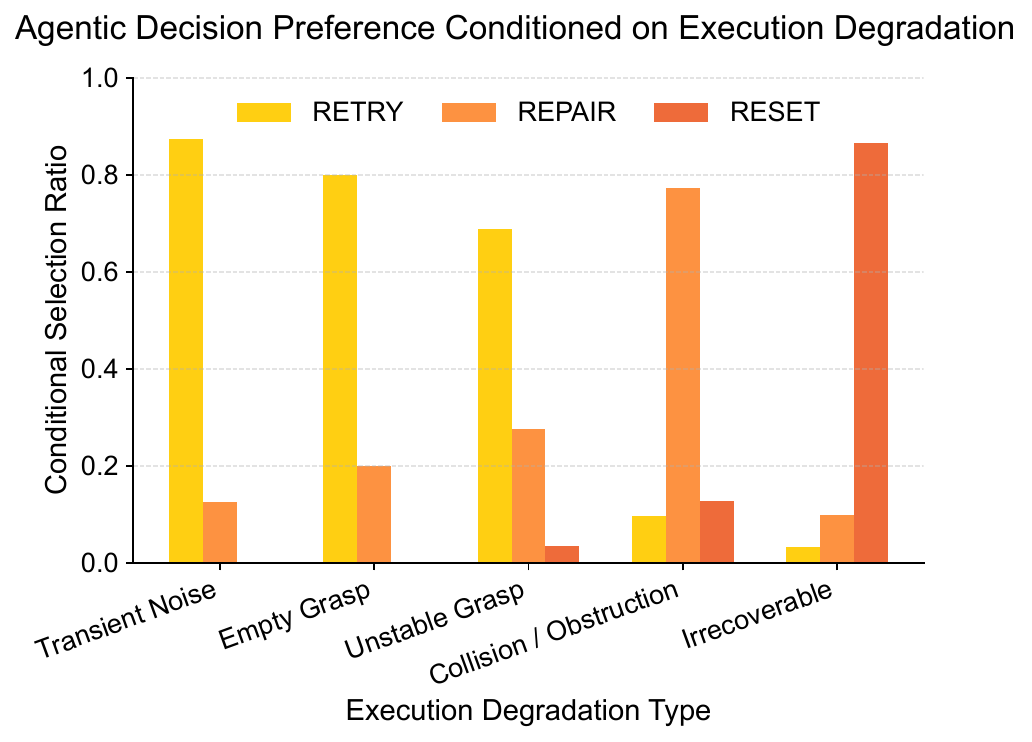}
    \caption{Conditional distribution of agentic decision across execution degradation types. Statistics are computed over trajectory segments where the agentic policy deviates from nominal execution. The agentic policy prefers \textsc{Retry} for mild degradations, \textsc{Repair} for collision-induced failures, and \textsc{Reset} for irrecoverable states.}
    \label{fig:dicision_distribution}
    \vspace{-11pt}
\end{figure}
We analyze the distribution of agentic decisions conditioned on execution degradation types to understand how the agentic policy adapts its behavior under execution degradation. The execution degradation types are defined based on observable execution-level characteristics, such as grasp outcomes, object motion, contact-induced constraints, and task feasibility. For this analysis, trajectory segments where the agentic policy deviates from nominal execution are manually inspected and categorized into degradation types according to these criteria. The statistics in Fig.~\ref{fig:dicision_distribution} are computed over trajectory segments where the agentic policy selects a non-\textsc{Execute} execution mode (\textsc{Retry}, \textsc{Repair}, or \textsc{Reset}).

As shown in Fig.~\ref{fig:dicision_distribution}, the agentic policy exhibits a preference pattern aligned with degradation severity. Mild degradations, such as transient noise and grasp-related failures, predominantly lead to \textsc{Retry} decisions, while collision or obstruction cases favor \textsc{Repair}. In irrecoverable states, such as when objects fall outside the reachable workspace or execution deviates from nominal behavior and execution quality remains persistently low, the policy tends to select \textsc{Reset}. This monotonic shift toward stronger execution modes indicates that the agentic policy differentiates execution degradation types and adapts its decisions accordingly, rather than relying on a single fixed recovery strategy.

\subsubsection{Agentic Decisions and Execution Quality (Q4)}
We analyze the relationship between agentic decisions and execution quality from two perspectives: how decision distributions vary with execution quality, and how different decisions affect subsequent execution quality.

\begin{figure}[t]
    \centering
    \includegraphics[width=0.6\columnwidth]{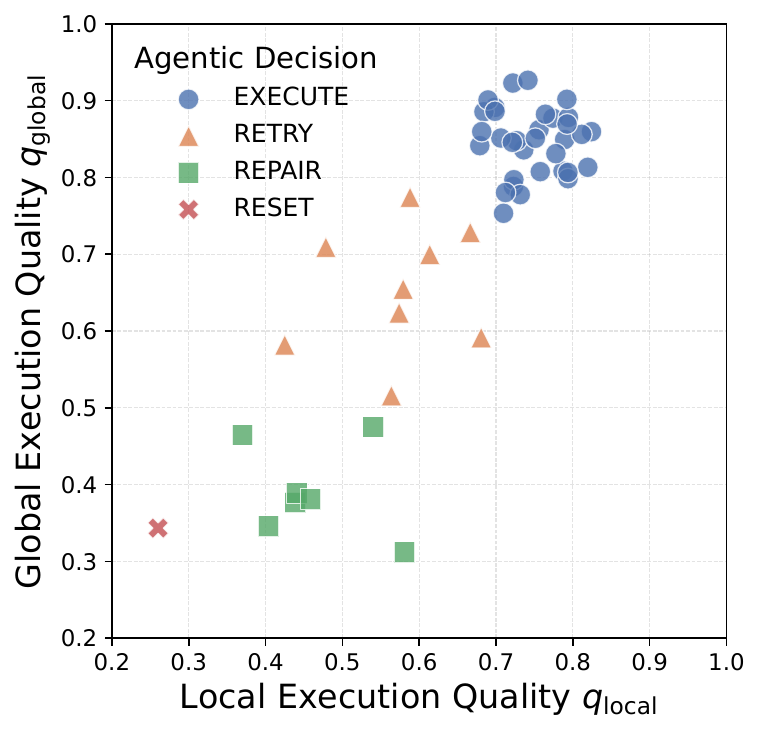}
    \caption{Agentic decisions plotted in the local–global execution quality space, showing how different execution modes are selected under varying execution conditions.}
    \label{fig:scatter}
    \vspace{-10pt}
\end{figure}

As shown in Fig.~\ref{fig:scatter}, different execution modes occupy distinct regions in the local–global execution quality space. \textsc{Execute} is primarily selected when both local and global quality are high, indicating stable and effective execution. \textsc{Retry} appears at intermediate quality levels, where execution is mildly degraded but still recoverable. In contrast, \textsc{Repair} and \textsc{Reset} are concentrated in low-quality regions, corresponding to more severe or irrecoverable degradation.

Table~\ref{tab:decision_effect} shows that these decisions have different effects on subsequent execution quality. The quality change is computed as the difference between execution quality at the decision step and that observed a fixed number of steps later. Values are reported as mean $\pm$ standard. \textsc{Execute} is often followed by slight quality decay, reflecting continued execution under marginal instability. Recovery actions yield positive quality changes. In particular, stronger recovery actions lead to higher probabilities of global quality improvement, confirming that the agentic policy not only conditions its decisions on execution quality, but also selects execution modes that induce appropriate, graded recovery effects.

\begin{table}[htbp]
\centering
\caption{Event-conditioned change in execution quality after different agentic decisions.}
\small
\setlength{\tabcolsep}{4pt}
\begin{tabular}{lccc}
\toprule
\textbf{Decision} &
$\Delta q_{\text{local}}$ &
$\Delta q_{\text{global}}$ &
$P(\Delta q_{\text{global}}>0)$ \\
\midrule
\textsc{Execute} &
$-0.01 \pm 0.02$ &
$-0.02 \pm 0.03$ &
$0.18$ \\

\textsc{Retry} &
$+0.08 \pm 0.04$ &
$+0.02 \pm 0.03$ &
$0.61$ \\

\textsc{Repair} &
$+0.05 \pm 0.03$ &
$+0.09 \pm 0.06$ &
$0.73$ \\

\textsc{Reset} &
$+0.20 \pm 0.08$ &
$+0.22 \pm 0.07$ &
$0.95$ \\
\bottomrule
\end{tabular}
\vspace{-2pt}
\label{tab:decision_effect}
\end{table}

\subsubsection{Recovery Cost and Efficiency (Q5)}
We analyze the execution overhead introduced by agentic decision-making in terms of recovery frequency and episode length. As summarized in Table~\ref{tab:intervention_cost}, based on 50 successful trajectories collected for each baseline policy on each task subset, augmenting the baseline with the agentic policy results in a moderate increase in the number of non-\textsc{Execute} modes, accompanied by a corresponding increase in episode length.

Notably, the overhead scales with the fragility of the low-level policy. Stronger policies such as $\pi_0$ and $\pi_{0.5}$ incur fewer recovery actions and smaller episode length increases, whereas more error-prone policies such as OpenVLA and the diffusion policy trigger more frequent recovery actions. This reflects the agentic policy’s tendency to intervene only when execution degradation is detected, rather than applying recovery indiscriminately.

Overall, the added execution cost remains modest relative to the substantial gains in success rate and robustness reported above.

\begin{table}[t]
\centering
\caption{Recovery cost and execution efficiency introduced by the agentic policy.}
\small
\setlength{\tabcolsep}{5pt}
\begin{tabular}{lcc}
\toprule
\textbf{Baseline Policy} &
\textbf{Number of Recoveries} &
\textbf{Episode Length} \\
\midrule
OpenVLA &
$+1.8 \pm 0.6$ &
$+12\%$ \\
$\pi_0$ &
$+1.2 \pm 0.4$ &
$+8\%$ \\
$\pi_{0.5}$ &
$+0.9 \pm 0.3$ &
$+5\%$ \\
Diffusion Policy &
$+2.1 \pm 0.7$ &
$+15\%$ \\
\bottomrule
\end{tabular}
\label{tab:intervention_cost}
\vspace{-6pt}
\end{table}

\section{CONCLUSIONS}

We propose an agentic reinforcement learning framework that improves robustness in robotic manipulation by separating execution management from low-level action generation. A high-level agentic policy reasons over execution history and selects execution modes that regulate how execution proceeds, while keeping the low-level policy frozen. This execution-level abstraction enables effective recovery from execution degradation without retraining task-specific skills or introducing additional planning modules. Experiments on the LIBERO benchmark demonstrate consistent improvements in task success rates for pretrained VLAs, together with interpretable execution-level behaviors.

A major limitation of our method is its limited ability to recover from more severe execution degradation and out-of-distribution scenarios. Since the agentic policy relies only on proprioceptive signals and low-level actions without requiring privileged simulator information, the framework is highly amenable to sim-to-real deployment. Future work will focus on improving recovery under severe degradation and validating the approach in real-world settings.

\addtolength{\textheight}{-12cm}

\bibliographystyle{IEEEtran}
\bibliography{IEEEabrv,ref}

\end{document}